\documentclass[11pt]{article}

\usepackage[final]{acl}

\usepackage{times}
\usepackage{latexsym}
\usepackage{booktabs}

\usepackage[T1]{fontenc}

\usepackage[utf8]{inputenc}

\usepackage{microtype}

\usepackage{inconsolata}
\usepackage{tcolorbox}

\newtcolorbox{promptbox}[1][]{
  colback=white,
  colframe=gray!60,
  coltitle=white,
  colbacktitle=gray!60,
  fonttitle=\bfseries\large,
  title={Prompt},
  rounded corners,
  boxrule=0.5pt,
  arc=3mm,
  #1
}

\newtcolorbox{custompromptbox}[2][]{
  colback=white,
  colframe=gray!60,
  coltitle=white,
  colbacktitle=gray!60,
  fonttitle=\bfseries\large,
  title={#2},
  rounded corners,
  boxrule=0.5pt,
  arc=3mm,
  #1
}

\usepackage{graphicx}

\title{H-RAG at SemEval-2026 Task 8: Hierarchical Parent–Child Retrieval for Multi-Turn RAG Conversations}

\author{
\textbf{Passant Elchafei}$^{1}$, 
\textbf{Hossam Emam}$^{3}$, 
\textbf{Mohamed Alansary}$^{3}$,\\
\textbf{Monorama Swain}$^{1}$ and
\textbf{Markus Schedl}$^{1,2}$ \\[3pt]
$^1$Johannes Kepler University Linz, Institute of Computational Perception, Linz, Austria \\
$^2$Linz Institute of Technology, Artificial Intelligence Lab, Austria \\
$^{3}$Independent Researcher, Egypt \\
\texttt{\{passant.elchafei, monorama.swain, markus.schedl\}@jku.at} \\
\texttt{\{hossamsalahemam, mohamed.alansary.c\}@gmail.com} \\
}

\begin{document}
\maketitle
\begin{abstract}
We present H-RAG, our submission to SemEval-2026 Task 8 (MTRAGEval), addressing both Task A (Retrieval) and Task C (Generation with Retrieved Passages). Task A evaluates standalone retrieval quality, while Task C assesses end-to-end retrieval-augmented generation (RAG) in multi-turn conversational settings, requiring both accurate answer generation and faithful grounding in retrieved evidence. Our approach implements a hierarchical parent--child RAG pipeline that separates fine-grained child-level retrieval from parent-level context reconstruction during generation. Documents are segmented into overlapping sentence-based child chunks, while full documents are preserved as parent units to provide coherent context. Retrieval combines hybrid dense--sparse search, tunable weighting, and embedding-based similarity rescoring over child chunks. Retrieved evidence is aggregated at the parent level and supplied to an instruction-tuned language model for response generation. H-RAG achieves an nDCG@5 score of 0.4271 on Task A and a harmonic mean score of 0.3241 on Task C (RB\_agg: 0.2488, RL\_F: 0.2703, RB\_llm: 0.6508), underscoring the importance of retrieval configuration and parent-level aggregation in multi-turn RAG performance.
\end{abstract}

\begin{figure*}[t]
  \centering
  \includegraphics[width=\textwidth]{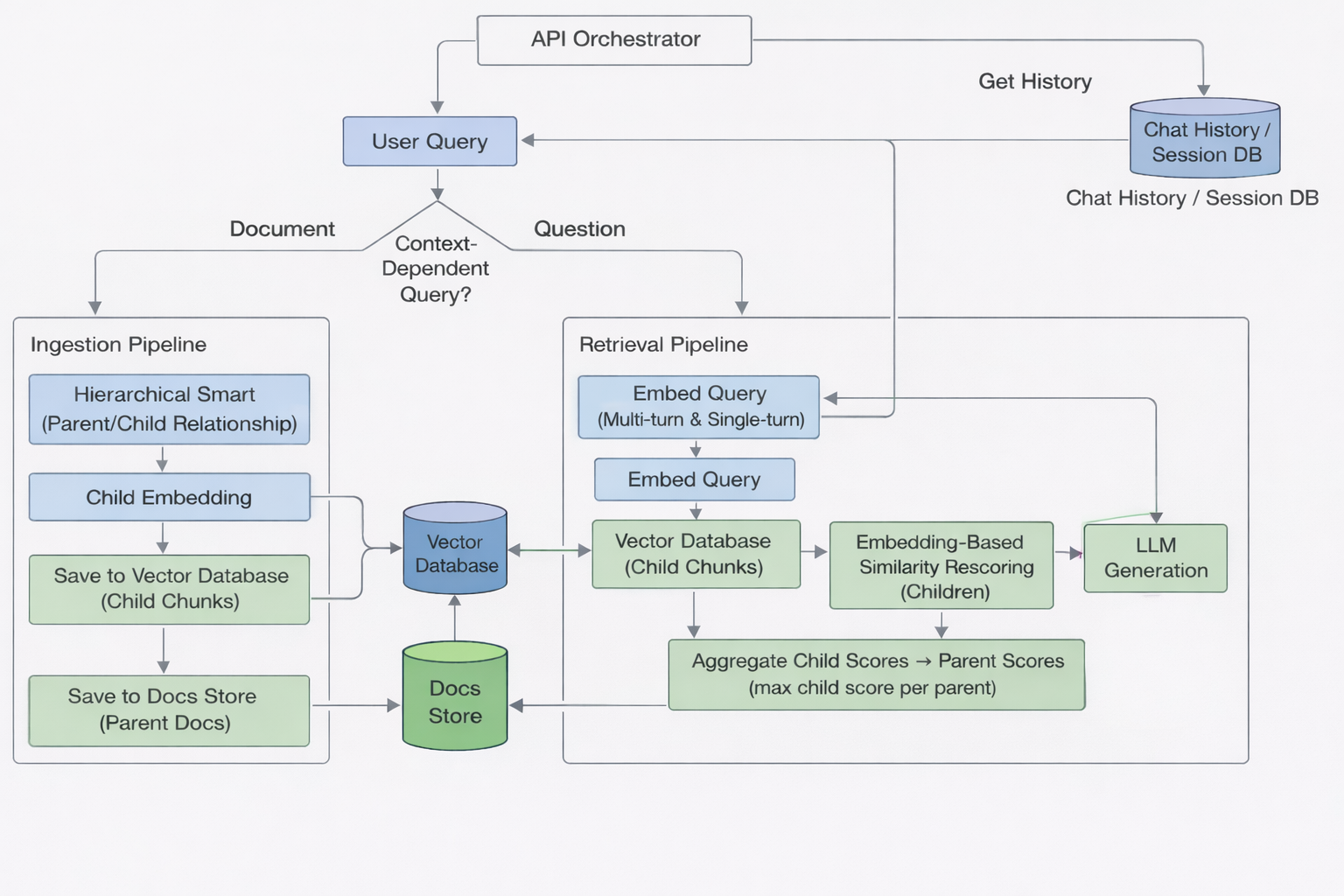}
  \caption{Overview of the proposed H-RAG pipeline for Task C, illustrating hierarchical parent–child document ingestion, hybrid retrieval over child chunks, embedding-based similarity rescoring, parent-level aggregation, and LLM-based answer generation in multi-turn conversational settings.}
  \label{fig:pipeline}
\end{figure*}

\section{Introduction}
Retrieval-Augmented Generation (RAG) has emerged as a standard paradigm for grounding large language model outputs in external knowledge sources \cite{lewis2020rag}. By retrieving relevant documents at inference time, RAG aims to improve factual accuracy and reduce hallucinations in knowledge-intensive tasks. While single-turn RAG pipelines have demonstrated strong performance, multi-turn conversational settings introduce additional challenges. User queries may depend on earlier dialogue turns, contain implicit references, or evolve in intent over time. These characteristics significantly increase the difficulty of both retrieving relevant evidence and maintaining faithful, grounded generation.

SemEval-2026 Task 8, MTRAGEval \cite{Rosenthal2026MTRAGEval} provides a comprehensive benchmark for evaluating multi-turn RAG systems under realistic conversational conditions \cite{katsis2025mtrag}. The shared task includes Task A (Retrieval), which evaluates standalone retrieval performance, and Task~C (Generation with Retrieved Passages), which assesses full end-to-end RAG. Task~C jointly measures answer quality, document-level grounding, and faithfulness, reflecting practical deployment requirements for RAG systems.

In this work, we present \textbf{H-RAG}, our system submission participating in both Task~A and Task~C. H-RAG is motivated by a fundamental design trade-off in RAG systems: fine-grained text chunks improve retrieval precision, whereas larger contextual units provide more coherent input for generation. To balance these competing objectives, we adopt a \textit{hierarchical parent--child representation that decouples retrieval granularity from generation context reconstruction}. Unlike prior hierarchical retrieval systems such as HRR \citep{singh2025hrr} and HiRAG \citep{huang2025hirag}, which introduce learned reranking or graph-structured knowledge integration, our contribution lies in the integrated design and configuration study of an unsupervised parent--child retrieval pipeline tailored to the multi-turn MTRAGEval setting. We further analyze how retrieval-side design choices influence end-to-end grounding and answer quality.

During document ingestion, each document is segmented into overlapping sentence-based child chunks optimized for semantic and lexical retrieval. In parallel, the original documents are preserved as parent units to provide coherent context once relevant evidence has been identified. At query time, the system optionally applies lightweight query rewriting for context-dependent conversational queries. Retrieval is performed using hybrid dense--sparse search with a tunable weighting parameter controlling the balance between embedding similarity and lexical matching \cite{cuconasu2024powerofnoise}. Retrieved child chunks are refined using embedding-based similarity rescoring, and evidence is aggregated at the parent level by assigning each parent document the maximum score of its associated children \cite{singh2025hrr, huang2025hirag}. The top-ranked parent documents are then supplied to an instruction-tuned language model for response generation.

\section{Related Work}

\paragraph{Multi-turn Conversational RAG.}
Recent work has emphasized evaluating retrieval-augmented generation (RAG) systems in multi-turn conversational settings, where dialogue context, topic shifts, and underspecified queries increase retrieval complexity. MTRAG \cite{katsis2025mtrag} introduces a human-authored benchmark for conversational retrieval and grounded generation. CORAL \cite{cheng2025coral} and RAD-Bench \cite{kuo2024radbench} further demonstrate the challenges of retrieval and reasoning across dialogue turns. These benchmarks highlight that conversational RAG performance depends heavily on robust cross-turn evidence selection. Our system is designed for this setting and evaluated under the MTRAGEval protocol.

\paragraph{Hybrid Dense--Sparse Retrieval.}
Hybrid retrieval combining dense embeddings and lexical matching has become a strong baseline in RAG systems. The Power of Noise study \cite{cuconasu2024powerofnoise} demonstrates that combining sparse and dense embeddings improves robustness over single-retriever approaches. DAT \cite{hsu2025dat} proposes query-adaptive weighting of dense and sparse scores. In contrast, H-RAG adopts a lightweight hybrid strategy with a statically tuned interpolation parameter $\alpha$, optimized via grid search to balance effectiveness and simplicity.

\paragraph{Hierarchical and Parent--Child Retrieval.}
Chunk granularity in RAG presents a recurring trade-off between retrieval precision and contextual coherence available to the generator. HRR \citep{singh2025hrr} proposes a learned hierarchical reranking framework over document trees, while HiRAG \citep{huang2025hirag} introduces graph-structured knowledge integration to aggregate multi-granular evidence. Both methods add supervised or structure-aware components on top of the base retriever. H-RAG follows the same intuition of combining fine-grained retrieval with higher-level context reconstruction, but adopts a lightweight unsupervised design. Sentence-level child chunks are retrieved and rescored using embedding similarity, then aggregated to parent documents via maximum-score assignment. Rather than proposing a new retrieval algorithm, our contribution is to study how hierarchical retrieval behaves under the MTRAGEval conversational benchmark and how retrieval-side configuration choices affect grounding quality.

\paragraph{Conversational Query Rewriting and Evaluation.}
Conversational retrieval often benefits from query rewriting to resolve context dependency. Multi-query rewriting \cite{kostric2024multiquery} and adaptive rewriting methods such as AdaQR \cite{zhang2024adaqr} improve recall by reformulating dialogue queries. H-RAG incorporates a lightweight rewriting step without introducing additional learned components.

Evaluation of RAG systems has evolved beyond traditional IR metrics. eRAG \cite{salemi2024erag} evaluates retrieval quality based on downstream utility, while RAGChecker \cite{ru2024ragchecker} provides fine-grained diagnostic analysis of retriever and generator contributions. \citet{wallat2025correctnessfaithfulness} demonstrate that citation correctness does not necessarily imply faithfulness. The MTRAG benchmark \cite{katsis2025mtrag} integrates these considerations into a multi-dimensional evaluation framework, under which we analyze both retrieval effectiveness (Task A) and grounded generation quality (Task C).

\section{System Overview}
\label{sec:system}

Our system, \textbf{H-RAG}, implements an end-to-end Retrieval-Augmented Generation (RAG) pipeline designed for Task C of the MTRAGEval shared task, where systems are required to retrieve relevant documents and generate grounded responses in multi-turn conversational settings. The architecture is designed to balance retrieval precision, contextual coherence, and computational efficiency, while remaining modular and scalable.

The pipeline consists of two tightly coupled phases: \emph{document ingestion} and \emph{query-time retrieval and generation}, as illustrated in Figure~\ref{fig:pipeline}.

\subsection{Document Ingestion}
\label{sec:ingestion}

During ingestion, documents are processed using a \textbf{hierarchical parent--child chunking strategy}. Each document is first segmented into sentences using a rule-based sentence splitter. Overlapping \emph{child chunks} are then constructed using a sliding window over consecutive sentences. In our implementation, each child chunk consists of three sentences with a stride of two sentences, resulting in overlapping chunks that preserve local coherence while remaining compact for retrieval.

Each child chunk is embedded using a sentence embedding model and indexed in a vector database to support efficient semantic similarity search. In parallel, the original full document text is preserved as a \emph{parent document} in a document store, along with metadata linking each child chunk to its corresponding parent. This hierarchical representation decouples retrieval granularity from generation context size: retrieval operates over fine-grained child chunks, while generation benefits from the richer context provided by parent documents.

\subsection{Query-Time Retrieval}
\label{sec:retrieval}

At query time, incoming user queries are handled by an API orchestrator that also retrieves recent dialogue history from a session store to support multi-turn interactions. When queries depend on prior conversational context or contain implicit references, a lightweight \textbf{LLM-based query rewriting module} reformulates them into standalone queries suitable for retrieval.

The rewritten query is embedded and used to perform hybrid dense--sparse retrieval over the indexed child chunks. Dense semantic similarity search is combined with keyword-based lexical matching using a tunable weighting parameter, allowing the system to balance semantic recall and exact-match precision. An initial set of top-$k$ child chunks is retrieved, where $k$ is treated as a configurable hyperparameter. The rewrite prompt instructs the model to return the question unchanged when it is already standalone; we found this rule essential for short factoid turns where LLM-generated rewrites tended to over-elaborate.

Retrieved child chunks are then refined using embedding-based similarity rescoring, where cosine similarity between the query embedding and child chunk embeddings is computed to improve ranking precision. Unlike cross-encoder reranking approaches, this rescoring step relies solely on embedding similarity, enabling efficient large-scale retrieval while maintaining ranking quality.

\subsection{Parent Aggregation and Context Reconstruction}
\label{sec:aggregation}

Following rescoring, child chunks are mapped back to their corresponding parent documents. Parent-level relevance scores are computed by aggregating the scores of associated child chunks, using the maximum child score as the parent score. The system supports two ranking strategies: (1) ranking child chunks first and then selecting unique parent documents based on aggregated scores, or (2) identifying unique candidate parent documents and performing an additional embedding-based rescoring step directly on parent document content. These strategies are treated as configurable options during experimentation.

The top-$n$ ranked parent documents are retrieved from the document store and used as context for generation. Supplying parent-level documents ensures that the generation module has access to sufficient surrounding information while avoiding excessive input length and redundancy.

\subsection{Answer Generation}
\label{sec:generation}

Finally, an instruction-tuned large language model generates a response conditioned on the reconstructed parent-level context and the rewritten query. The generation module is designed to either produce a grounded answer supported by the retrieved documents or abstain when the available evidence is insufficient, in accordance with Task C requirements. Overall, H-RAG integrates hierarchical document representation, hybrid retrieval, embedding-based rescoring, and parent-level context reconstruction into a unified pipeline for multi-turn RAG. This design aligns closely with the MTRAGEval evaluation framework, which assesses both answer quality and document-level evidence grounding.

\subsection{Implementation Details}
\label{sec:impl}
Documents are segmented into overlapping three-sentence child chunks with a stride of two sentences using a rule-based sentence splitter. Child chunks and parent documents are indexed in a Weaviate hybrid vector store with $\alpha = 0.7$ (70\% dense, 30\% sparse). Child embeddings are produced using \texttt{BAAI/bge-large-en-v1.5}, while rescoring uses \texttt{BAAI/bge-reranker-v2-m3} through cosine similarity over embeddings.

The initial candidate set size is $k = 50$ child chunks. Parent scores are computed using maximum-score aggregation over retrieved child chunks, and the top-$n = 5$ parent documents are passed to the generator. Both query rewriting and answer generation use OpenAI \texttt{gpt-5}. Query rewriting uses temperature $0.2$ to minimize unnecessary reformulation, while answer generation uses temperature $0.7$ with \texttt{max\_completion\_tokens = 4096}. No component is fine-tuned; all models are used zero-shot. The generation prompt instructs the model to respond conversationally while avoiding explicit references to ``the context'' or numeric citations. The full prompt templates are provided in Appendix~\ref{app:prompts}. Additional configuration ablation results are provided in Appendix~\ref{app:ablation}.
Table~\ref{tab:settings} summarizes the implementation settings used for the submitted H-RAG system, including embedding model, hybrid retrieval configuration, chunking strategy, ranking method, and LLM parameters.
\section{Experimental Setup}
\label{sec:experiments}

\subsection{Dataset}
\label{sec:dataset}

We evaluate our system on the MTRAG dataset released as part of SemEval-2026 Task 8: MTRAGEval. The dataset consists of human-curated multi-turn conversational interactions grounded in a large document collection. Each instance contains a sequence of dialogue turns followed by a final user query, which may depend on earlier conversational context through coreference, ellipsis, or implicit constraints.

In \textbf{Task C: Generation with Retrieved Passages}, systems are required to retrieve a fixed number of relevant documents from the corpus and generate a grounded response conditioned on the retrieved evidence. The dataset includes both answerable and unanswerable queries, requiring systems to either produce a faithful answer or abstain when insufficient evidence is available. We use the official dataset splits and evaluation protocol provided by the task organizers.

\subsection{Retrieval Configuration and Hyperparameters}
\label{sec:retrieval_config}

Our experiments focus on analyzing the impact of key retrieval and ranking components within the H-RAG pipeline. Rather than introducing additional architectural components, we perform a systematic exploration of retrieval-related hyperparameters that directly influence evidence selection and grounding quality. We explore the following configuration dimensions:

\paragraph{Embedding Model.}
We evaluate two sentence embedding models for indexing and retrieval: \texttt{intfloat/multilingual-e5-large} and \texttt{bge-large-en-v1.5}. These models are used consistently for child chunk embedding and similarity computation during retrieval and rescoring.

\paragraph{Hybrid Retrieval Weight.}
Hybrid dense--sparse retrieval combines semantic similarity search with keyword-based lexical matching. We tune the hybrid weighting parameter $\alpha$, which controls the relative contribution of dense embeddings versus sparse keyword matching. We consider values $\alpha \in \{0.9, 0.7, 0.5\}$.

\paragraph{Ranking Strategy.}
We evaluate two parent document ranking strategies. In the first strategy, child chunks are ranked and rescored first, after which unique parent documents are selected based on aggregated child scores. In the second strategy, unique parent documents are identified earlier in the pipeline and subjected to an additional embedding-based rescoring step using parent-level content. This setting allows us to assess the effect of ranking order on evidence selection.

\paragraph{Initial Candidate Set Size.}
We vary the number of retrieved child chunks prior to parent aggregation. Specifically, we consider candidate set sizes of $k \in \{20, 30, 50\}$, enabling analysis of the trade-off between retrieval recall and ranking noise. Unless otherwise specified, a single configuration is designated as the base setup for controlled comparisons, and remaining parameters are held constant while individual dimensions are varied.

\subsection{Evaluation Protocol}
\label{sec:evaluation}

We evaluate all configurations using the official MTRAGEval evaluation framework. The framework combines reference-based and reference-less metrics designed specifically for RAG systems. Evaluation assesses answer quality, document-level evidence grounding, and appropriate abstention behavior for unanswerable queries. All system outputs follow the official Task C submission schema, which requires generated responses to be accompanied by a ranked list of retrieved documents. Scores are computed automatically using the task-provided scripts, and all evaluations are conducted under identical conditions.

\begin{table}[t]
\centering
\small
\begin{tabular}{l l}
\toprule
Component & Name \\
\midrule
Embedding model & BAAI/bge-large-en-v1.5 \\
Reranker & BAAI/bge-reranker-v2-m3 \\
Generation LLM & GPT-5 (T=0.7) \\
Query rewrite LLM & GPT-5 (T=0.2) \\
Hybrid weight $\alpha$ & 0.7 \\
Initial candidates $k$ & 50 \\
Top-$n$ parents & 5 \\
Chunking & 3 sentences, stride 2 \\
Aggregation & Max child score \\
Ranking strategy & Child-first \\
Vector store & Weaviate hybrid \\
\bottomrule
\end{tabular}
\caption{Implementation settings used for the submitted H-RAG system.}
\label{tab:settings}
\end{table}

\section{Results and Analysis}
\label{sec:results}

We report results using the official MTRAGEval evaluation framework. Task A evaluates standalone retrieval quality using nDCG@5, while Task C evaluates end-to-end RAG using the harmonic mean of document relevance (RB\_agg), faithfulness (RL\_F), and LLM-based answer quality (RB\_llm). The reported scores correspond to the configuration using \texttt{BAAI/bge-large-en-v1.5}, $\alpha = 0.7$, ranking by child-first aggregation, and an initial candidate size of $k=50$. 

Table~\ref{tab:ablation} summarizes the configuration sweep across $\alpha \in \{0.5, 0.7, 0.9\}$, $k \in \{20, 30, 50\}$, and both ranking strategies on the development set. Ranking strategy emerges as the dominant factor: enabling parent-level rescoring (\texttt{rank\_parents = true}) consistently improves retrieval metrics across all configurations, with mean gains of +0.0197 nDCG@5 and +0.0108 Recall@5. In contrast, hybrid weighting has limited sensitivity, with less than 0.003 variation in mean nDCG@5 across $\alpha$ values. Candidate size peaks at $k=30$, suggesting that larger retrieval pools may introduce additional ranking noise. The strongest configuration is achieved with $\alpha = 0.7$, parent-level ranking enabled, and $k=30$, yielding nDCG@5 = 0.4872 and Recall@5 = 0.5184. Additional configuration results across hybrid weighting, ranking strategy, and candidate set size are reported in Appendix~\ref{app:ablation}.

\subsection{Task A: Retrieval Performance}

On Task A, H-RAG achieves an nDCG@5 score of \textbf{0.4271} on the final submission dataset. It further achieves an nDCG@5 score of \textbf{0.4728} and Recall@5 of \textbf{0.502} on the baseline dataset \cite{Rosenthal2026MTRAGEval}, while the baseline model reports an nDCG@5 score of \textbf{0.45} and Recall@5 of \textbf{0.49}. These results reflect the effectiveness of hybrid dense--sparse retrieval combined with hierarchical child-level indexing. Sentence-level child chunks improve fine-grained matching, while hybrid fusion balances lexical precision and semantic similarity. Overall, retrieval design choices strongly influence performance in multi-turn conversational settings.

After comparing document-level and passage-level datasets \cite{rosenthal2026mtragunbenchmarkopenchallenges}, we observed an approximately 10\% improvement when evaluating on document-level data. This gain suggests that the proposed system is particularly well suited for long-reference-document scenarios, aligning with its design objective of supporting short-query, long-context retrieval settings.

Compared with baseline results reported by the task organizers \cite{rosenthal2026mtragunbenchmarkopenchallenges, Rosenthal2026MTRAGEval}, H-RAG achieves slight improvements, demonstrating its effectiveness and robustness relative to established approaches.

\subsection{Task C: End-to-End RAG Performance}

For Task C, H-RAG achieves a harmonic mean score of \textbf{0.3241}, with component scores of RB\_agg = 0.2488, RL\_F = 0.2703, and RB\_llm = 0.6508. The relatively high RB\_llm score suggests that the generation module produces fluent and semantically appropriate responses. However, lower RB\_agg and RL\_F scores indicate that evidence grounding remains the primary bottleneck. The gap between RB\_llm and grounding-related metrics suggests that retrieval quality alone does not guarantee faithful evidence usage during generation. While retrieved parent documents often contained relevant supporting information, the generation model occasionally produced fluent responses that were only partially aligned with retrieved evidence. This indicates that evidence selection and grounding remain more challenging than answer fluency in conversational RAG.

\section{Conclusion}
We presented H-RAG, a hierarchical parent--child retrieval system developed for SemEval-2026 Task~8 (MTRAGEval), participating in both Task~A and Task~C. The proposed architecture decouples fine-grained child-level retrieval from parent-level context reconstruction, integrating hybrid dense--sparse search, embedding-based rescoring, and lightweight query rewriting within a unified pipeline. Our results demonstrate that, in multi-turn conversational RAG, fluent answer generation alone is insufficient for strong end-to-end performance; robust evidence selection and aggregation remain dominant limiting factors. The observed gap between answer quality and grounding metrics highlights the role of retrieval design in conversational settings. Future work may explore adaptive hybrid weighting, refined parent-level aggregation, uncertainty-aware retrieval mechanisms, context-aware segmentation, and conversational-history summarization to strengthen grounded generation in long-context dialogue systems.

\bibliography{custom}
\appendix
\section{Prompt Templates}
\label{app:prompts}
\begin{custompromptbox}{Query Rewrite Prompt}
\setlength{\fboxsep}{1pt}
\small

\textcolor{teal}{\textbf{Task:}} 
\textit{Given the conversation history, rewrite the new user question into a standalone and specific query suitable for retrieval.}

\vspace{0.3em}

\textcolor{blue}{\textbf{Important Rules:}}
\begin{itemize}
\setlength{\itemsep}{1pt}
\setlength{\parskip}{0pt}
\setlength{\parsep}{0pt}
\setlength{\topsep}{2pt}
    \item If the question is already clear and standalone, return it \textbf{EXACTLY} as is
    \item If the question contains pronouns or references to earlier dialogue, rewrite it using the necessary context
    \item Do NOT invent information or change the original meaning
    \item Return only the rewritten question with \textbf{NO explanation}
\end{itemize}

\vspace{0.3em}

\textcolor{teal}{\textbf{Conversation History:}} \\
\texttt{\{history\_text\}} \\
\textit{(last three Q/A turns, formatted as ``Q: ...'' and ``A: ...'')}

\vspace{0.3em}

\textcolor{blue}{\textbf{New Question:}} \texttt{\{user\_question\}}

\vspace{0.3em}

\textcolor{purple}{\textbf{Expected Output:}} \\
\textit{Rewritten standalone question (or unchanged original question if already standalone).}
\end{custompromptbox}
\begin{custompromptbox}{Generation Prompt}
\setlength{\fboxsep}{3pt}
\small
\textcolor{teal}{\textbf{Task:}} 
\textit{Generate a natural conversational response grounded in retrieved parent-level documents while maintaining dialogue continuity.}
\vspace{0.3em}
\textcolor{blue}{\textbf{System Instruction:}} You are a helpful AI assistant engaged in a conversation with a user. Answer the user's question naturally and directly, as if you already know the information.
\vspace{0.3em}
\textcolor{blue}{\textbf{Important Rules:}}
\begin{itemize}
\setlength{\itemsep}{1pt}
\setlength{\parskip}{0pt}
\setlength{\parsep}{0pt}
\setlength{\topsep}{2pt}
    \item Do NOT mention ``the context'', ``the provided information'', ``according to the documents'', or similar phrases
    \item Do NOT reference source numbers such as [1], [2], etc.
    \item Respond conversationally as if the knowledge is your own
    \item If insufficient information is available, state this naturally without mentioning missing context
    \item Maintain continuity with the conversation history
\end{itemize}
\vspace{0.3em}

\textcolor{teal}{\textbf{Conversation History:}} \\
\texttt{[Conversation history block, when present]}

\vspace{0.3em}

\textcolor{teal}{\textbf{Retrieved Background Knowledge:}} \\
\texttt{[Top-$n$ parent documents with numbered references]} \\
\textit{(used internally for grounding; never referenced explicitly)}

\vspace{0.3em}

\textcolor{blue}{\textbf{Current User Question:}} \texttt{\{user\_question\}}

\vspace{0.3em}

\textcolor{purple}{\textbf{Generation Trigger:}} \\
\texttt{Respond naturally and conversationally:}

\end{custompromptbox}

\section{Configuration Ablation}
\label{app:ablation}
\begin{table*}[t]
\centering
\small
\begin{tabular}{c c c c c c c c c}
\toprule
$\alpha$ & rank\_parents & $k$ & nDCG@1 & nDCG@3 & nDCG@5 & Recall@1 & Recall@3 & Recall@5 \\
\midrule
0.5 & true  & 20 & 0.4938 & 0.4530 & 0.4832 & 0.2111 & 0.4243 & 0.5120 \\
0.5 & true  & 30 & 0.4952 & 0.4511 & 0.4826 & 0.2120 & 0.4204 & 0.5108 \\
0.5 & true  & 50 & 0.4931 & 0.4449 & 0.4765 & 0.2093 & 0.4105 & 0.5046 \\

0.7 & true  & 20 & 0.4913 & 0.4513 & 0.4822 & 0.2092 & 0.4236 & 0.5125 \\
0.7 & true  & 30 & \textbf{0.4952} & \textbf{0.4559} & \textbf{0.4872} & \textbf{0.2111} & \textbf{0.4288} & \textbf{0.5184} \\
0.7 & true  & 50 & 0.4938 & 0.4506 & 0.4806 & 0.2110 & 0.4204 & 0.5084 \\

0.9 & true  & 20 & 0.4913 & 0.4515 & 0.4828 & 0.2095 & 0.4227 & 0.5125 \\
0.9 & true  & 30 & 0.4938 & 0.4562 & 0.4856 & 0.2104 & 0.4296 & 0.5145 \\
0.9 & true  & 50 & 0.4926 & 0.4537 & 0.4852 & 0.2113 & 0.4258 & 0.5164 \\

\midrule

0.5 & false & 20 & 0.4425 & 0.4248 & 0.4618 & 0.1910 & 0.4068 & 0.5022 \\
0.5 & false & 30 & 0.4373 & 0.4245 & 0.4622 & 0.1884 & 0.4076 & 0.5048 \\
0.5 & false & 50 & 0.4464 & 0.4285 & 0.4648 & 0.1931 & 0.4105 & 0.5046 \\

0.7 & false & 20 & 0.4514 & 0.4310 & 0.4653 & 0.1944 & 0.4115 & 0.5029 \\
0.7 & false & 30 & 0.4438 & 0.4291 & 0.4657 & 0.1916 & 0.4115 & 0.5064 \\
0.7 & false & 50 & 0.4476 & 0.4316 & 0.4662 & 0.1934 & 0.4139 & 0.5050 \\

0.9 & false & 20 & 0.4360 & 0.4153 & 0.4433 & 0.1878 & 0.3967 & 0.4746 \\
0.9 & false & 30 & 0.4502 & 0.4305 & 0.4673 & 0.1942 & 0.4117 & 0.5069 \\
0.9 & false & 50 & 0.4361 & 0.4250 & 0.4624 & 0.1880 & 0.4095 & 0.5058 \\
\bottomrule
\end{tabular}
\caption{Configuration ablation across hybrid weighting $\alpha$, ranking strategy, and initial candidate size $k$. Best-performing configuration is highlighted in bold.}
\label{tab:ablation}
\end{table*}

Table~\ref{tab:ablation} reports the full configuration sweep across hybrid weighting $\alpha$, ranking strategy, and candidate set size $k$. These experiments were conducted to analyze retrieval sensitivity under the MTRAGEval framework.

\end{document}